\pdfoutput=1

\documentclass[11pt]{article}

\usepackage[]{EMNLP2023}

\usepackage{times}
\usepackage{latexsym}
\usepackage{graphicx}

\usepackage[T1]{fontenc}

\usepackage[utf8]{inputenc}

\usepackage{microtype}

\usepackage{inconsolata}

\title{Ranking LLMs by compression}

\author{ Peijia Guo$^{1,2}$ , Ziguang Li $^{3}$ , Haibo Hu $^{3}$ , Chao Huang $^{3*}$, Ming Li $^{4*}$ , Rui Zhang $^{1*}$\\
        $^{1}$ School of Mathematics, Northwest University, Xi’an, China \\$^{2}$Shanghai Institute for Mathematics and Interdisciplinary Sciences, Shanghai, China
        \\$^{3}$ Institute of Computing Technology, Chinese Academy of Sciences, China \\$^{4}$Cheriton School of Computer Science, University of Waterloo, Ontario, Canada\\ Guopeijia0929@163.com~~~~~~chriszggz@gamil.com~~~~~~huhaibo22@mails.ucas.ac.cn~~~~~~\\chuang@ict.ac.cn~~~~~~mli@uwaterloo.ca~~~~~~rzhang@nwu.edu.cn}

\begin{document}
\maketitle
\begin{abstract}
We conceptualize the process of understanding as information compression, and propose a method for ranking large language models~(LLMs) based on lossless data compression. We demonstrate the equivalence of compression length under arithmetic coding with cumulative negative log probabilities when using a large language model as a prior, that is, the pre-training phase of the model is essentially the process of learning the optimal coding length. At the same time, the evaluation metric compression ratio can be obtained without actual compression, which greatly saves overhead. In this paper, we use five large language models as priors for compression, then compare their performance on challenging natural language processing tasks, including sentence completion, question answering, and coreference resolution. Experimental results show that compression ratio and model performance are positively correlated, so it can be used as a general metric to evaluate large language models.

\end{abstract}

\section{Introduction}
In recent years, the rapid development of LLMs has brought earth-shaking changes to the field of natural language processing (NLP) (\citealp{radford2019language};~\citealp{zhao2023survey};~\citealp{liu2023summary}). LLMs are advanced language models pretrained on tens of gigabytes of data without tuning on data for specific tasks. These large models can directly complete various NLP tasks, and even become a milestone technology towards general artificial intelligence (AGI). Currently, LLMs are being studied more and more widely in various fields, such as education and research~\citep{app13095783}, medicine and healthcare~(\citealp{thirunavukarasu2023large};~\citealp{cascella2023evaluating}), etc., and their performance evaluation methods are becoming more and more important. 

\citet{chang2024survey} showed that researchers always scrutinize the capabilities of AI models or algorithms through evaluation using specific and challenging tasks, so the evaluation metrics are outlined from the perspective of the evaluation tasks. The metrics are diverse, such as Exact Match (EM), F1-score, ROUGE, etc., and many are set for specific tasks, making it difficult to uniformly evaluate the performance of the model on different tasks. In addition, contamination of training and test data can also lead to biased evaluation results~\citep{magar2022data}, making it impossible to verify whether NLP progress is achieved through better language understanding or better data utilization. Various limitations lead to the lack of a unified LLMs evaluation standard. 

Therefore, we consider the process of model training and learning itself and prove the equivalence of the model pre-training goal and the compression length under arithmetic coding, indicating that compression is closely related to model performance, and then use the compression ratio as a general metric to measure the model's generalization ability in different scenarios. 

\section{Related Work}

\subsection{Language Models Evaluation}

Currently, performance evaluation of LLMs is mainly achieved through benchmark tests, including diverse tasks, standardized datasets and comprehensive evaluation metrics. The purpose is to establish a systematic and standardized evaluation framework.

In 2019,~\citet{wang2019glue}~introduced the General Language Understanding Evaluation Benchmark (GLUE), a multi-task evaluation platform for measuring the performance of natural language understanding models. It contains nine tasks, covering various types such as text classification, text similarity evaluation, natural language Inference, question answering, etc. A recent study \citet{laskar2023systematic}~evaluated ChatGPT across 140 tasks and analyze 255K responses it generates in these datasets, laying the foundation for deploying ChatGPT-like LLMs in real-world applications. More recently, \citet{openai2024gpt4} tested GPT-4 on a diverse set of benchmarks, including 34 simulating exams that were originally designed for humans. Benchmark test is very important for evaluating the performance of language models and promoting research progress, but limited coverage tasks, data contamination~(\citealp{brown2020language};~\citealp{li2023open}), and huge overhead are all challenges and limitations faced in this process. In order to solve these problems, we propose compression ratio based on lossless data compression, a general evaluation metric.

\subsection{Neural Compression}

The goal of data compression is to reduce the representation size while retaining valid information. Our LLMs-based compressor uses neural networks for data compression and belongs to the neural compression category. Current research in neural compression largely benefits from advances in deep generative modeling~\citep{yang2023introduction}, such as GANs~\citep{goodfellow2014generative}, VAEs~\citep{rezende2015variational}, and autoregressive models~\citep{van2016pixel}. With the development of deep neural networks, lossless text compression has also ushered in new progress. \citet{goyal2018deepzip}~ proposed DeepZip, a lossless compressor based on neural networks, consisting of two main modules: RNN and arithmetic coding. It achieves higher compression ratio than GZIP. ~\citet{bellard2019lossless}~ proposed a lossless compressor based on LSTM, which is simple to describe and has reasonable memory consumption compared to compressors that provide a similar compression ratio. Recent advancements, such as TRACE, a fast transformer-based general-purpose lossless compressor~\citep{mao2022trace}, achieves an overall speedup of approximately 3x while maintaining a compression ratio comparable to state-of-the-art compressors.

\section{Method}

\subsection{LLMs based Arithmetic Coding for Compression}

 Shannon’s fundamental theorem of coding states that~\citep{shannon1948mathematical}, given messages randomly generated from a model, it is impossible to encode them into less bits (on average) than the entropy of that model, thus defining a lower bound for lossless compression. Arithmetic coding is an entropy coding algorithm. \citet{huang2023approximating}~proposed an entropy-based compressor that integrated generative pre-trained transformer into
adaptive arithmetic coding, highlighting the potential of pre-trained LLMs as powerful priors in compression. In this paper, we integrate LLMs into adaptive arithmetic coding for compression, with the aim of representing data according to the probability of output to reduce its overall size.

\textbf{LLMs as Entropy Models}

Considering text data, first use a tokenizer to convert the text into a data stream $t_{1:n}:=t_{1}t_{2}\cdots t_{n}\in T^{n}$ of length $n$, where $T$ is LLM vocabulary, a finite set of tokens. The empty sequence is denoted as $\varepsilon$. Let $\phi$ represents LLM, where $\phi(t_{1:(i-1)})= P_{i}(t_{i}|t_{1},t_{2},\cdots , t_{i-1}),~i\ge2$ means modeling the next token $t_{i}$ through the previous $i-1$ tokens $t_{1:(i-1)}$, and we get its probability distribution $P_{i}$. In order to obtain the distribution for $P_{1}$, add an EOS (End of Sentence) token at the beginning of the text as $t_{0}$. For each token $t_{i}$, the associated $P_{i}$ acts as the entropy model, guiding the encoder to allocate fewer bits to high-frequency tokens and more bits to low-frequency tokens, thereby improving compression efficiency.

\textbf{Coding Process}

The range for the data stream is the interval $\left [ 0,1\right ) $ before anything is transmitted. As each token is processed, the cumulative distribution functions $F_{i}(t_{i})$ and $P_{i}(t_{i})$ are calculated according to $\phi(t_{0:(i-1)})$. Then narrow the interval to the part assigned to that token: $$I_{low}^{i}=I_{low}^{i-1}+(I_{high}^{i-1}-I_{low}^{i-1})*F_{i}(t_{i}),$$  
$$I_{high}^{i}=I_{low}^{i-1}+(I_{high}^{i-1}-I_{low}^{i-1})*(F_{i}(t_{i})+P_{i}(t_{i}))$$

\begin{table}
\centering
\begin{tabular}{lc}
\hline
\textbf{Algorithm 1} Arithmetic Coding\\
\hline
1: \textbf{Input}:~$t_{0:n}:=t_{0}t_{1}\cdots t_{n}\in T^{n+1}$. \\
2:~$I_{low}^{0}=0,~I_{high}^{0}=1 $\\
3: \textbf{for}~ $t_{i},~i=1,2,\cdots,n,n+1$~ ~\textbf{do}\\
4:~~~~~$range=I_{high}^{i-1}-I_{low}^{i-1}$\\
5:~~~~~$I_{low}^{i} \gets I_{low}^{i-1} +range*F_{i}(t_{i})$\\ 
6:~~~~~$I_{high}^{i} \gets I_{low}^{i-1} +range*(F_{i}(t_{i}) + P_{i}(t_{i}))$\\
7: \textbf{end for}\\ 
8: \textbf{Output}:  $[I_{low}^{n+1},I_{high}^{n+1})$.\\\hline
\end{tabular}
\label{tab:accents}
\end{table}

Adaptive arithmetic coding using LLM is shown in Algorithm 1.

\subsection{Equivalence of Model Pre-training Goal and Compression Length}

It is well established that compression and prediction are essentially equivalent ~\citep{deletang2023language}. In this way, compression and LLMs are closely linked. We mathematically prove the equivalence of model pre-training goal and compression length. Then we present a novel method for evaluating LLMS based on lossless compression.

\textbf{Pre-training Optimization Goals for LLMs}

The loss function, also known as the objective function, measures the difference between the probability distribution predicted by the model and the true distribution. Model training is to reduce the loss function through continuous iteration, thereby optimizing model performance.

We continue to consider the data stream above $t_{1:n}:=t_{1}t_{2}\cdots t_{n}\in T^{n}$. The true distribution of data $Q$ is the sequence of probability mass functions $Q_{n}:T^{n}\to(0, 1]$, for all $n\in N$, satisfying the constraint $Q_{n}(t_{1:n})= {\textstyle \sum_{s\in T}^{}Q_{n+1}(t_{1:n}s)} $, where $Q_{0}(\varepsilon ):=1$. The meaning can be clearly seen from the parameters of $Q$, so we omit the subscript of $Q$.

Now we have the true distribution $Q$ of the data and the probability distribution $P$ predicted by the LLMs. The pre-training Optimization Goals for LLMs is to make $P$ closer to $Q$, which can elicit the definition of relative entropy, that is, Kullback-Leibler Divergence:$$D_{KL}(Q||P)={\sum_{i=1}^{n}(Q_{i} \log_{2}{Q_{i}})}-{ \sum_{i=1}^{n}(Q_{i} \log_{2}{P_{i}})}$$

The previous term ${\sum_{i=1}^{n}(Q_{i}\cdot \log_{2}{Q_{i}})}$ is the inverse of the entropy of the true distribution $Q$, which is constant. The last term $-{ \sum_{i=1}^{n}(Q_{i}\cdot \log_{2}{P_{i}})}$ is the definition of cross entropy, represented by $H (Q,P)$. Gibbs inequality states that~\citep{gibbs1878equilibrium}:  $D_{KL}(Q||P)\ge 0$, the equality sign is true if and only if $Q_{i}=P_{i},\forall i$. Therefore, in order to make the probability distribution $P$ closer to the true distribution $Q$, that is, to minimize the value of cross entropy. It further illustrates that cross entropy can be used as the loss function, and minimizing cross entropy is the goal of optimizing the model.

\textbf{Negative Log Probability as Compression Length}

The goal of lossless compression is to encode a data stream $t_{1:n}$ sampled from a true distribution $Q$ into a minimum length bit stream, while ensuring that the original sequence can be recovered through decoding. In practice, $Q$ is usually unknown, so we approximate $Q$ through the probability distribution $P$ predicted by the LLMs $\phi$. During arithmetic coding, the length of the interval $I^{i}$ is equal to $I^{i-1}*P_{i}(t_{i})$. For the sequence $t_{1:n}$, starting from the initial interval of length 1, the final encoded interval length is ${\textstyle \prod_{i=1}^{n}} P_{i}(t_{i })$, so the number of bits required to represent this final interval (i.e. message $t_{1:n}$) is $\textstyle \sum_{i=1}^{n}-log_{2}P_{i}(t_{i}) $. This reveals a direct way to approximate the compression length without having to perform the compression method exactly. So the expected number of bits we get is $E_{t\sim Q}\left [ {\textstyle \sum_{i=1}^{n}-\log_{2}{P_{i}(t_{i}})} \right ]$, that is the cross entropy $H(Q,P)$.

Therefore, in the process of achieving lossless compression, minimizing the expected length of the encoded data stream is equivalent to minimizing cross entropy. At this point, the equivalence of model pre-training goal and compression length has been proven. Furthermore, we can use compression ratio as a unified criterion for evaluating LLMs.

\section{Experiments}

The experiment consists of four key parts: the calculation of the compression ratio and three natural language processing tasks, namely sentence completion, question answering and coreference resolution. We use a total of five LLMs as compressor priors, but the proposed method is not limited to these models. This method can be applied to more advanced LLMs as long as the predicted probabilities can be obtained.

\subsection{The Calculation of Compression Ratio}

First, we select the Text8 dataset to calculate the compression ratio of the compressor. The Text8 dataset is a large corpus extracted from the English Wikipedia. After some simple preprocessing, the text content covers various topics and fields. It is a general dataset for language modeling.

We split the read Text8 file by spaces and obtain a list containing all words. Then every 200 words are divided into a sublist, and the 200-length word fragment are converted into strings. The list of the first 10,000 strings is passed to the LLMs compressor as a parameter. The compression ratio calculation formula is as follows (in bits):$$compression~ratio = \frac{original~text~length}{compressed~text~length}.$$

The LLM compressors involved include LLaMA~2~7B released by Meta, Mistral~7B released by the Mistral AI team, OPT-IML~1.3B released by Facebook, and GPT-2-XL~1.5B and GPT-2~774M released by OpenAI. Their calculated compression ratios are shown in Table 1.

\begin{table}
\centering
\begin{tabular}{lc}
\hline
\textbf{Compressor} & \textbf{Compression Ratio} \\
\hline
LLaMA~2~7B & 8.663 \\
Mistral~7B & 9.266 \\
OPT-IML~1.3B & 6.938\\
GPT-2-XL 1.5B & 7.095 \\
GPT-2 774M    & 6.864 \\
\hline
\end{tabular}
\caption{\label{citation-guide}
Compression ratios of different compressors.
}
\end{table}

\subsection{Sentence Completion}

Sentence completion is designed to allow the computer to predict the missing parts based on the given context, so that the sentence becomes coherent and complete. We compare the performance of three large models, LLaMA~2~7B, Mistral~7B and GPT-2-XL~1.5B on the HellaSwag dataset, using accuracy as a metric. The results are shown in Table 2.

\begin{table}
\centering
\begin{tabular}{lc}
\hline
\textbf{LLM} & \textbf{Accuracy(\%)} \\
\hline
Mistral 7B & 81.3 \citep{jiang2023mistral}\\
LLaMA 2 7B~& 77.2 \citep{touvron2023llama}\\
GPT-2-XL 1.5B & 50.9 \citep{wu2023lamini}\\
\hline
\end{tabular}
\caption{\label{citation-guide}
Performance on sentence completion .
}
\end{table}

\subsection{Question Answering}

The goal of question answering is to enable the computer to understand the questions raised by users through semantic understanding and syntax analysis, and then generate answers that meet the requirements of the questions. Because any form of LLM evaluation can be seen as question answering or switch to this format, so it is a very important means for LLMs evaluation\citep{guo2023evaluating}. We compare the performance of two large models, LLaMA~2~7B and OPT-IML~1.3B on the BoolQ dataset, using accuracy as a metric. The results are shown in Table 3.

\begin{table}
\centering
\begin{tabular}{lc}
\hline
\textbf{LLM} & \textbf{Accuracy(\%)} \\
\hline
LLaMA 2 7B & 77.4 \citep{touvron2023llama}\\
OPT-IML 1.3B & 61.5 \citep{iyer2023optiml}\\
\hline
\end{tabular}
\caption{\label{citation-guide}
Performance on question answering.
}
\end{table}

\subsection{Coreference Resolution}

Coreference resolution is to identify the entities referred to by pronouns and noun phrases in the text. It has many practical applications in natural language processing, such as information extraction, text summarization, etc. Correct parsing of reference relationships can help computers better understand text. We compares the performance of two large models, GPT-2-XL~1.5B and GPT-2~774M on the Winograd Schema Challenge data set, using accuracy as a metric. The results are shown in Table 4.
\begin{table}
\centering
\begin{tabular}{lc}
\hline
\textbf{LLM} & \textbf{Accuracy(\%)} \\
\hline
GPT-2-XL 1.5B & 73.3 \citep{wu2023lamini}\\
GPT-2 774M    & 69.2 \citep{trichelair2018reasonable}\\
\hline
\end{tabular}
\caption{\label{citation-guide}
Performance on coreference resolution.
}
\end{table}

\subsection{Result Analysis}

From the above experiments, it can be concluded that: the better data compression effect of LLM, the better its performance in natural language processing tasks. That is, there is a positive correlation between compression ratio and model performance.

When we can effectively compress data, it means that we have captured the key characteristics and patterns of the data. This is similar to finding patterns and redundancies in the data during the model learning process. So we can say that if a large language model achieves the best lossless compression on a dataset, it will often achieve the best generalization on other datasets.

Therefore, the experimental results further verify the theoretical conclusion of this paper: compression ratio can be used as a general metric to measure the performance of LLMs.

\section{Conclusion}

We proposed to rank LLMs through
lossless data compression in this paper. Our method measures compression ratios as a metric for generalization. We demonstrate the equivalence of compression length under arithmetic coding and LLMs pre-training goal, saving the overhead of actual compression. This further illustrates that understanding is compression, demonstrated by our experiments across challenging downstream NLP tasks.

\section{Limitations}

For NLP tasks, the experiments in this paper only used the open source version of the pre-trained language model, which was subject to computational constraints and scale limitations. Furthermore evaluation is not the end goal but the starting point. A mature evaluation system should not only provide conclusions about performance, but also provide analysis and guidance for future research and development, which is also our future research direction.

\section{Statement}

We take academic integrity and research independence very seriously. Here we would like to declare that parts of this paper overlap with a published paper. 
Overlaps include ideas presented, experimental methods. Information about the published paper is as follows:

\begin{itemize}
\item Title: Compression Represents Intelligence Linearly
\item Author: Yuzhen Huang, Jinghan Zhang, Zifei Shan, Junxian He
\item arXiv:2404.09937 [cs.CL]
\item Submission date: April 15, 2024
\end{itemize}

When we began our work, we were unaware of the existence of this published paper. Our study began on December 2023, was completed on May 2024, and was submitted on June 20, 2024.

\end{document}